## Acknowledgments

Preprint. Under review.

# Pareto-Enhanced Portrait Generation: Vision-Aligned Text Supervision for Alignment, Realism, and Aesthetics

Yunlong Wang
Xiaomi Corporation
Nanjing, China

Jinjin Shi*
Xiaomi Corporation
Nanjing, China

Wenbin Gao
Xiaomi Corporation
Nanjing, China

Xuran Xu
Xiaomi Corporation
Nanjing, China

Runyu Shi
Xiaomi Corporation
Beijing, China

Ying Huang
Xiaomi Corporation
Beijing, China

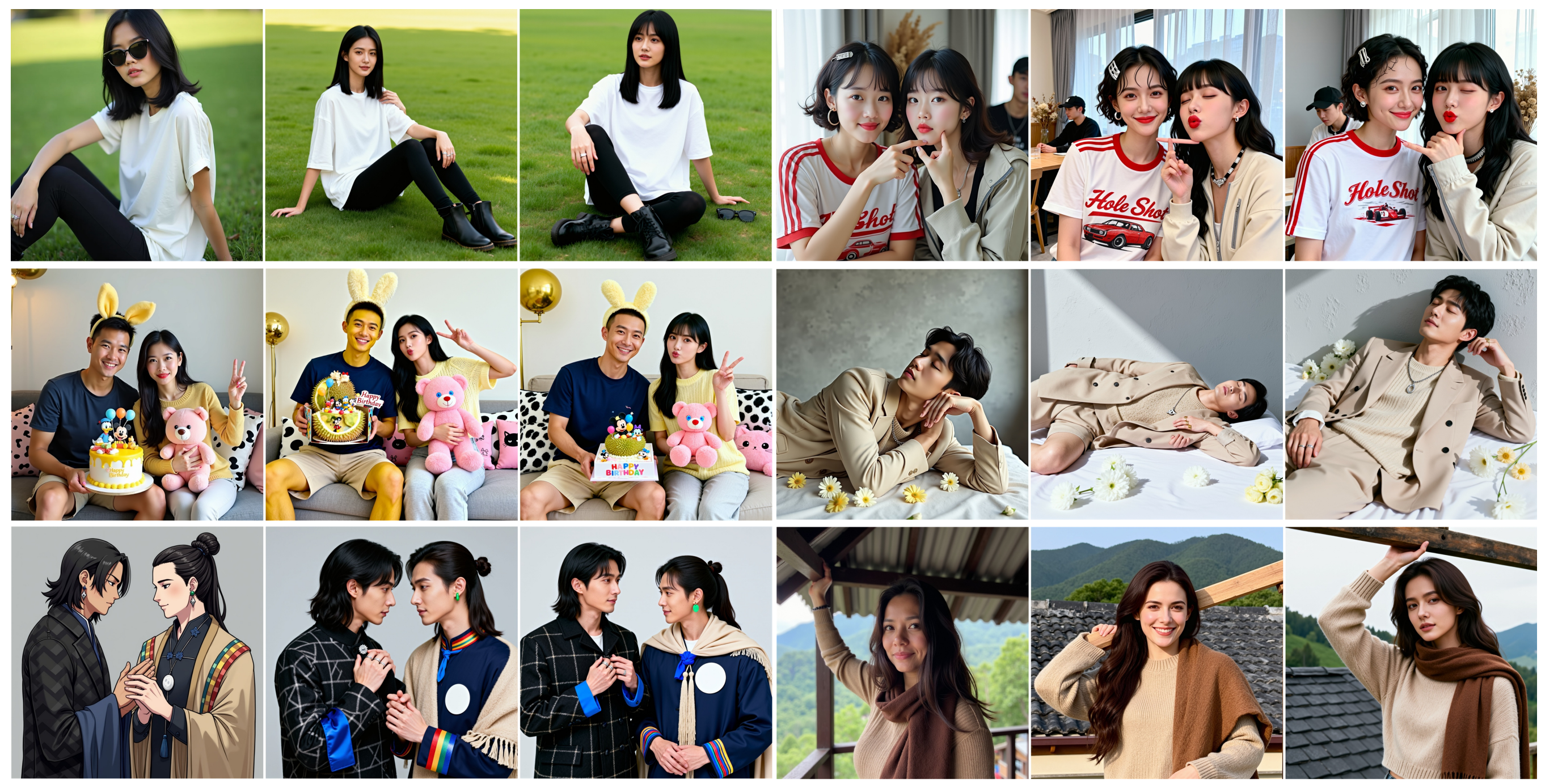

**Figure 1: Comparison of generated human images across three settings: Flux.1-dev baseline (left), baseline + SFT (middle), and baseline + ours (right). Compared to the baseline and SFT counterparts, our method achieves the best performance in terms of photorealism, human aesthetics, and text-image alignment, thereby achieving synergistic improvements across all three dimensions.**

## Abstract

Text-to-image diffusion models often face a severe trilemma in human portrait generation: text-image alignment, photorealism, and human-perceived aesthetics inherently inhibit one another. Supervised Fine-Tuning (SFT) is an effective method for enhancing the photorealism of image generation. However, it often leads to overfitting to the training dataset, corrupts pre-trained image priors, and degrades alignment or aesthetics. To break this bottleneck, we propose a feature supervision paradigm for Multimodal Diffusion Transformers (MM-DiT). Specifically, we introduce a lightweight cross-modal alignment mechanism that implicitly extracts multi-granularity vision-aligned text representations from SigLIP 2 and applies supervision to the image branch of MM-DiT during the training stage, with zero extra inference overhead. Our method injects vision-aligned text guidance while preserving the base model's original generalization, avoiding degradation caused by SFT. Furthermore, our method directly mines implicit multi-granularity aesthetic signals from pre-trained vision foundation models to optimize human-perceived aesthetics. Extensive experiments on MM-DiTs show that our method pushes the Pareto frontier and achieves synergistic improvements across text-image alignment, photorealism, and human-perceived aesthetics.

*Corresponding author.

## Keywords

Text-to-Image Generation, Diffusion Models, MM-DiTs, Human Portrait Generation, Feature Supervision, Pareto Optimality

## 1 Introduction

Human image generation is a crucial aspect of computer vision, owing to its wide range of applications in digital content creation, immersive virtual reality experiences, personalized social media avatar customization, and advanced cinematic post-production, among other industrial uses. Recent advancements in diffusion probabilistic models [3, 6, 20, 22, 23] have significantly boosted contemporary text-to-image synthesis technologies. Leading models like Stable Diffusion [20], SDXL [17], DALL·E 3 [1], especially Flux [9][1], which is based on MM-DiT [30], show remarkable capability to transform varied textual descriptions into visually captivating images, excelling in high-resolution rendering and overall visual quality.

Nevertheless, despite these notable technical strides, human image generation, which can be considered the most important but perception-sensitive aspect within text-to-image synthesis, continues to face three major challenges. These fundamental bottlenecks significantly limit the reliability and practical application of existing models in professional and industrial scenarios:

- **Inadequate Human-Perceived Aesthetics**: The human cognitive system possesses a highly specialized sensitivity to the aesthetic harmony and visual pleasantness of human portraits. However, existing foundational models frequently struggle to align with these human-perceived aesthetic standards [28, 29]. They often produce visually discordant or uninspiring appearances. Even trivial aesthetic deviations in facial lighting or discordant color tones can instantly disrupt the subjective visual harmony.
- **Insufficient Photorealism**: While modern diffusion models can generate visually plausible human images at a coarse semantic level, they consistently fail to capture the subtle and high-frequency details that define real human portraits, such as organic skin textures and natural specular gloss, a gap that directly restricts their application in professional production workflows demanding ultra-high visual quality.
- **The Trade-Off among Text Alignment, Photorealism and Human-Perceived Aesthetics**: We observe a persistent Pareto frontier [14] among text alignment, photorealism, and human-perceived aesthetics, where maximizing one metric inherently degrades the others. For instance, forcing the model to memorize photorealistic details traps the network in severe overfitting, causing the model's output to collapse into a rigid pixel domain. Consequently, this leads to a severe suppression of comprehensive human-perceived aesthetics, and potentially results in degraded text alignment performance.

To address these limitations, the research community has explored three mainstream technical trajectories in recent years. However, each method is plagued by inherent drawbacks that severely restrict its practicality, scalability, and overall performance:

*Supervised Fine-Tuning.* To address the aforementioned realism gap, a straightforward strategy is direct SFT on real-world, high-quality human datasets (e.g., Majicmix Realistic [13]). While aggressively updating parameters successfully optimizes high-realism visual content and local details, this intensive optimization comes at a severe cost. It typically traps the network in overfitting, significantly degrades its overall cross-modal generative adaptability, and compromises its human-perceived aesthetics.

*DreamBooth Tuning.* Subject-driven personalization techniques, prominently represented by DreamBooth [21], excel at binding unique text identifiers to specific subjects using minimal reference images. To mitigate overfitting during fine-tuning, DreamBooth explicitly employs a class-specific prior preservation loss. Nevertheless, these methods are intrinsically tailored for instance-level memorization rather than domain-wide distribution alignment. They necessitate computationally prohibitive per-subject optimization and are thus unscalable for general high-realism portrait synthesis.

*Reinforcement Learning.* Preference alignment methods including DPO [19] and SPO [33] aim to align model outputs with human aesthetic preferences, but they incur prohibitive data collection and computational costs. Specifically, these methods necessitate either training highly complex and often unstable visual reward models, or constructing massive, manually annotated paired-preference datasets [8, 28]. Moreover, the final generation quality is strictly upper-bounded by the scale and reliability of the preference data, severely limiting their scalable adaptation across diverse domains.

To break the fundamental trade-off among text alignment, photorealism, and human-perceived aesthetics, we propose a novel, parameter-efficient framework that integrates superior vision-aligned semantic features from an advanced Vision Foundation Model (VFM [32]), i.e., SigLIP 2 [24], into the MM-DiT architecture, as shown in Figure 2.

Despite the technical breakthroughs of MM-DiT architectures in enabling unified token-level interactions between text and images, they inherently suffer from fundamental limitations in generating high-perceptual-quality human portraits under the conventional SFT paradigm. This deficiency stems not from a lack of architectural expressivity, but from a severe mismatch between the global supervision signals provided by SFT and the intrinsic demand for precise patch-level semantic alignment in human aesthetics. Human-perceived aesthetics are exceptionally demanding, they rely not merely on global semantic correctness, but heavily on the localized aesthetic attributes of specific image patches. Specifically, while the self-attention mechanism of MM-DiT structurally permits dense token-to-patch interactions, SFT relies exclusively on macroscopic latent-based Flow Matching loss. This singular reconstruction objective completely lacks explicit patch-level image-semantic constraints.

[1]This paper adopts FLUX.1-dev (available at https://huggingface.co/black-forest-labs/FLUX.1-dev), which is licensed under the FLUX.1 [dev] Non-Commercial License v1.1.1. The authors confirm that all uses of the aforementioned model in this work is limited to academic research purposes only, and no commercial activities are involved.

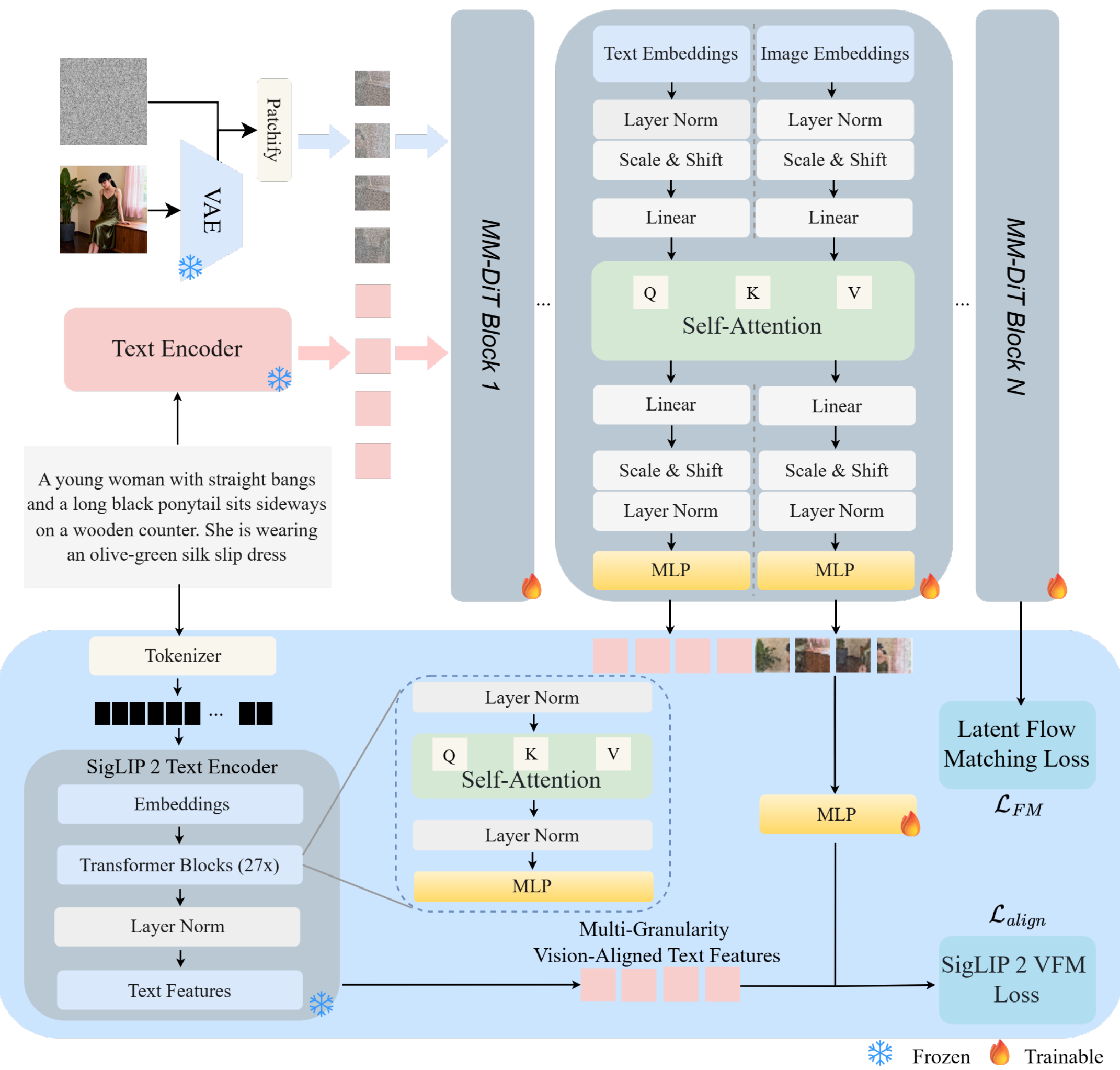

**Figure 2: An overview of our proposed SigLIP 2 vision-aligned text feature supervision method. SigLIP 2 text feature supervision module is marked inside the red dashed box. Beside the basic flow matching loss, we construct a supervision loss between the vision-aligned SigLIP 2 text features and the image features at the depth-n layer of MM-DiT.**

Empowered by LocCa [26] and SILC/TIPS [24] training objectives, SigLIP 2 inherently provides multi-granularity cross-modal priors for the generative process. Specifically, LocCa enforces patch-level cross-modal alignment between fine-grained text descriptions and local image regions, endowing the encoder with dense spatial perception and precise local semantic binding. Furthermore, the SILC self-distillation mechanism leverages an EMA teacher to regularize local-to-global feature consistency, effectively capturing both high-level semantics and delicate structural details. Concurrently, the TIPS masked prediction objective drives the encoder to learn intricate local spatial dependencies and preserve high-frequency aesthetic details by reconstructing masked patches. Together, these complementary training paradigms empower SigLIP 2 to generate highly reliable, multi-granularity visual features.

Owing to this rigorous cross-modal optimization, SigLIP 2's textual features are inherently mapped to a highly expressive visual space. Consequently, we propose to utilize SigLIP 2's vision-aligned text embeddings as the direct supervisory signal for the MM-DiT image branch. This paradigm offers two advantages: on one hand, these vision-aligned text features act as a rich proxy for complex micro-aesthetic details and high-fidelity structural priors; on the other hand, since the supervisory signal natively originates from the text modality, it intrinsically tightens the text-image binding, thereby synergistically boosting prompt adherence and breaking the traditional zero-sum bottleneck. We present a visual overview of our main results in Figure 1. In the right column, our method eliminates skin tone bias and body structure collapse arising from SFT. Moreover, our method achieves highly accurate text-image alignment. As shown in the first row, our results strictly adhere to the complex prompts "sunglasses at the feet" and "a girl closing one eye" whereas both the baseline and SFT versions fail to capture these details.

In summary, the core contributions of this work are three-fold:

*(1) A Non-intrusive and Parameter-Efficient Supervision Paradigm:* We construct a lightweight supervision framework that directly refines the MM-DiT based model using features extracted from SigLIP 2, without introducing cumbersome auxiliary modules or expanding model parameters. This paradigm significantly elevates photorealism and human-perceived aesthetics while preserving the base model's original open-domain generalization capabilities, effectively avoiding the degradation issues typical of conventional fine-tuning.

*(2) Synergistic Enhancement of Text Alignment, Photorealism and Human-Perceived Aesthetics:* Refining the inherent trade-off among text alignment, photorealism and human-perceived aesthetics in conventional fine-tuning, our method extracts vision-aligned text features from advanced VFMs, resolves the inherent zero-sum dilemma, ultimately generating visually striking, intricately detailed, and aesthetically aligned natural portraits while ensuring text alignment and faithful prompt adherence.

*(3) Cost-Free Human Aesthetics Improvement without Resource-Intensive Constraints:* We completely bypass the need for massive manually annotated paired data or the prohibitive training overhead of reward models required by conventional preference alignment pipelines like DPO or RLHF. Human-perceived aesthetics fundamentally rely on multi-granularity structural harmony. Our method directly mines and harnesses implicit multi-granularity priors from pre-trained VFMs, and optimizes human-perceived aesthetics with zero additional inference overhead.

## 2 Related Work

### 2.1 Image Generation with Diffusion Models

Diffusion probabilistic models have become the dominant architecture for high-realism visual content generation, thanks to their strong data distribution learning ability and progressive iterative denoising framework [6]. Latent Diffusion Models (LDM) [20] shift computational cost from high-dimensional pixel space to a compact compressed latent space, greatly improving large-scale generation efficiency. Following this route, Stable-Diffusion-v1.5 [20] uses a U-Net backbone with CLIP-based [18] text conditioning via cross-attention and a $512 \times 512$ base resolution, marking a milestone in efficiency and open ecosystem development. Constrained by its training resolution, it lacks fine high-frequency details. SDXL [17] adopts stronger text encoders and a two-stage base+refiner pipeline, lifting the base resolution to $1024 \times 1024$ and greatly enhancing photorealism. This improvement comes with higher inference overhead, and the model still suffers from the inherent limitations of its U-Net architecture, and still falls short on fine-grained text-image alignment. Stable Diffusion 3 [4] utilizes MM-DiT based joint-attention Transformer backbone, enabling dense text-image token interactions and boosting robustness for complex prompts and text rendering, yet it entails higher training and inference costs. FLUX combines latent flow matching with dual-stream MM-DiT and single-stream DiT, striking a better balance between speed, fidelity and prompt alignment, while the lack of explicit word-region supervision causes unstable gains in fine human details.

### 2.2 Human Image Generation

To enhance human photorealism, conventional practices often resort to direct Supervised Fine-Tuning (SFT) on high-quality human datasets. This method often triggers severe training set overfitting. Although aggressively updating parameters helps optimize high-realism visual details, it comes at the cost of degrading the model's cross-modal generative adaptability, compromising its original versatility and human-perceived aesthetics. For instance, Majicmix Realistic achieves impressive performance in bust portrait generation, yet it suffers from issues such as unstable full-body anatomical structures, monotonous facial styles, and insufficient generalization ability. Then subject-driven personalization techniques like DreamBooth are introduced to refine the human portraits synthesis. However, these methods necessitate computationally prohibitive per-subject optimization, making them fundamentally unscalable for general high-realism portrait synthesis.

Aligning generative model outputs with human-perceived aesthetic preferences and textual intent is a core challenge for practical real-world deployment. While Reinforcement Learning from Human Feedback (RLHF) [15] has become the prevailing standard for large language models, its direct application to diffusion models faces insurmountable hurdles: training a stable and accurate reward model for high-dimensional visual data is exceptionally difficult, and the associated computational cost is prohibitively high for large-scale training.

Direct Preference Optimization (DPO) has emerged as a popular alternative, optimizing the model directly on annotated preference pairs (winning vs. losing samples) without training an explicit reward model. Representative methods such as Diffusion-DPO [25] have achieved noticeable progress in human image quality. For instance, HelloWorldXL [11] is one of the most popular photorealistic portrait foundational models in the SDXL ecosystem improved by Diffusion-DPO. It exhibits outstanding advantages particularly in Asian portrait generation. However, these methods suffer from two critical bottlenecks that limit their scalability and performance upper bound:

- **Heavy Data Annotation Dependency**: These methods rely heavily on large-scale, manually curated preference datasets [8], which require massive labor costs and are difficult to adapt and expand to diverse human generation scenarios.
- **Synthetic Data-Driven Performance Ceiling**: Most existing DPO frameworks construct preference pairs using only synthetically generated images, trapping the model in a closed learning loop where performance is strictly capped by the quality of the initial synthetic data, and failing to fully leverage the distribution of real high-quality human images.

## 3 Method

### 3.1 Preliminaries

This section introduces the foundational theories and architectures of our work, including Latent Flow Matching (LFM) and Multimodal Diffusion Transformers (MM-DiT), which form the basis of modern text-to-image models.

*Latent Flow Matching.* Flow Matching [2, 12] is an efficient generative paradigm that learns a continuous ordinary differential equation (ODE) to map a simple noise distribution to a target data distribution. For image generation, a pre-trained variational autoencoder (VAE) [7] first compresses high-resolution images into a lower-dimensional latent space. Formally, LFM defines a continuous time-dependent vector field $v_\theta(z_t, t, c)$, where $z_t$ is the latent variable at time $t \in [0, 1]$, $c$ is the text condition, and $\theta$ denotes model parameters. The model is trained to fit the flow path from initial Gaussian noise $z_0$ to the clean latent $z_1$. The generation is achieved by solving the ODE:

$$\frac{dz_t}{dt} = v_\theta(z_t, t, c). \tag{1}$$

Compared to the Markov-based iterative denoising in traditional diffusion models, LFM provides straighter generation trajectories, enabling few-step fast generation with minimized cumulative errors.

*Multimodal Diffusion Transformers (MM-DiT).* MM-DiT advances standard Diffusion Transformers (DiT) [16] by introducing a dual-stream architecture. Instead of concatenating visual latents and textual conditions into a single sequence, MM-DiT treats them as parallel, independent token sequences. The modalities interact symmetrically via a joint attention mechanism before passing through modality-specific MLPs. With time-step and guidance signals injected through AdaLN, this deep multimodal joint attention overcomes the local receptive field limits of U-Net, significantly enhancing long-range dependencies and text-image alignment.

*MM-DiT based Text-to-Image Models.* FLUX is a state-of-the-art text-to-image generative model proposed by Black Forest Labs, built on the combination of latent flow matching and a hybrid transformer backbone comprising both dual-stream (MM-DiT) and single-stream (DiT) blocks, featuring targeted structural innovations to balance generation speed, visual fidelity, and text alignment. Alongside FLUX, QwenImage [27] represents another leading paradigm in this domain. Driven by robust native VLM encoders and optimized MM-DiT structures, QwenImage significantly enhances the model's instruction-following capabilities for highly complex prompts and provides exceptional native support for multilingual text-to-image synthesis, further expanding the application boundaries of MM-DiT architectures.

## 3.2 SigLIP 2 Vision-Aligned Text Feature Supervision

Our method is based on the Flow Matching framework, as depicted in Figure 2. Unlike stable diffusion models that predict noise via score matching, Flow Matching aims to regress a continuous vector field that transports a simple prior distribution to the complex data distribution.

Let $z_1 \sim p_{\text{data}}$ denote the latent representation of a real image, and $z_0 \sim \mathcal{N}(0, I)$ denote the initial Gaussian noise. The continuous time variable is defined as $t \in [0, 1]$. Following the Rectified Flow formulation, we construct a linear probability path interpolating between the noise and the data:

$$z_t = tz_1 + (1 - t)z_0. \tag{2}$$

The corresponding target marginal vector field (i.e., the velocity) that drives this linear trajectory is obtained by taking the time derivative of $z_t$:

$$u_t(z_t) = \frac{dz_t}{dt} = z_1 - z_0. \tag{3}$$

The generative model, parameterized by $\theta$ (typically a Diffusion Transformer), outputs a predicted velocity $v_\theta(z_t, t, c)$ conditioned on the timestep $t$ and text prompt $c$. The network is primarily optimized via a continuous-time Mean Squared Error (MSE) loss:

$$\mathcal{L}_{\text{FM}} = \mathbb{E}_{t \sim \mathcal{U}(0,1), z_0, z_1, c} \left[ \| v_\theta(z_t, t, c) - (z_1 - z_0) \|_2^2 \right]. \tag{4}$$

While $\mathcal{L}_{\text{FM}}$ enforces the macroscopic trajectory learning, generic text conditions often lack the capacity to supervise fine-grained portrait details. To explicitly inject vision-aligned text priors without corrupting the learned vector field, we introduce an auxiliary supervision signal guided by SigLIP 2. Let $F_{\text{siglip}}(c)$ denote the fine-grained text embeddings extracted by the frozen SigLIP 2 text encoder, and $H_{\text{img}}(z_t, \theta)$ denote the corresponding visual patch features extracted from the image branch of the MM-DiT, we define a semantic alignment loss by computing the distance between these two feature spaces:

$$\mathcal{L}_{\text{align}} = \mathbb{E}_{t, z_t, c} \left[ \mathcal{D} \left( F_{\text{siglip}}(c), \Phi(H_{\text{img}}(z_t, \theta)) \right) \right], \tag{5}$$

where $\Phi(\cdot)$ is a lightweight projection layer aligning the MM-DiT visual features to the SigLIP 2 textual semantic space, and $\mathcal{D}(\cdot, \cdot)$ represents the distance metric.

The overall training objective of our method is formulated as a weighted addition:

$$\mathcal{L}_{\text{total}} = \mathcal{L}_{\text{FM}} + \lambda \mathcal{L}_{\text{align}}, \tag{6}$$

where the hyperparameter $\lambda$ controls the supervision strength of the vision-aligned text features.

During inference, the supervision branch is discarded. Starting from a pure noise sample $z_0$, the generation process is formulated as solving an Ordinary Differential Equation (ODE): $z_1 = z_0 + \int_0^1 v_\theta(z_t, t, c)\, dt$, which yields the final generated latent code, i.e., the standard latent flow matching generation pipeline.

# 4 Experiments

In this section, we detail our experimental framework. We begin by introducing the experimental details, datasets, and evaluation protocols. Subsequently, we evaluate the effectiveness of our method on mainstream benchmarks, followed by ablation studies to assess the efficacy of each component. In our experiments, unless explicitly stated otherwise, the default configuration adopts SigLIP 2 supervision, and employs MLP for projection, with injection depth of 6 and weighting coefficient of 0.1.

## 4.1 Experimental Settings

*Datasets.* We collect an internal dataset consisting of roughly 200k high-quality human images to facilitate robust training. We randomly sample 5k real-world images from this collection as a held-out test set. The remaining data are used for training. Similar to standard practices, we apply filtering to ensure data quality.

**Table 1: Comparison of our method with SFT on Flux.1-dev and Qwen-Image baselines.**

| Method | FID ↓ | CLIPScore(%) ↑ | HPSv2.1(%) ↑ | ImageReward(%) ↑ |
|---|---|---|---|---|
| Flux.1-dev (Baseline) | 35.17 | 79.36 | 30.50 | 21.32 |
| + SFT | 27.94 | 81.25 | 30.49 | 18.60 |
| + ours | **27.40** | **81.40** | **30.54** | **21.57** |
| Qwen-Image | 33.30 | 79.77 | 29.64 | 11.05 |
| + SFT | 28.16 | 82.53 | 30.37 | 21.51 |
| + ours | **28.13** | **83.04** | **30.44** | **24.88** |

**Table 2: Comparison of our method with Flux.1-Krea-dev on the Flux.1-dev baseline.**

| Method | FID ↓ | CLIPScore(%) ↑ | HPSv2.1(%) ↑ | ImageReward(%) ↑ |
|---|---|---|---|---|
| Baseline | 35.17 | 79.36 | 30.50 | 21.32 |
| + ours | **27.40** | **81.40** | **30.54** | **21.57** |
| Flux.1-Krea-dev | 28.20 | 79.90 | 30.08 | 15.14 |

*Metrics.* To evaluate the generation performance, we employ four quantitative metrics focusing on distribution realism, human-perceived aesthetics, and text-image alignment: 1) **Image Realism**: We adopt Fréchet Inception Distance(FID) [31] to measure the distance between the distribution of generated images and real images to evaluate realism and diversity. 2) **Human Aesthetic Preference Metrics**: We utilize ImageReward [29] to assess the generated images' alignment with human aesthetics and plausibility. This metric is trained on extensive human preference datasets. 3) **Text-Image Alignment**: We use CLIPScore [5] and HPSv2 [28] to quantify the semantic consistency between the generated images and the input text prompts. Note that for FID, a lower value indicates better performance, whereas for HPSv2, ImageReward, and CLIPScore, higher values denote superior quality.

*Baselines.* We implement our training method on two distinct models to validate its generality: 1) Flux.1-dev: A state-of-the-art open-weight text-to-image model developed by Black Forest Labs. Built upon a hybrid architecture that integrates MM-DiT with Flow Matching. 2) Qwen-Image: An open-source image generation foundation model developed by Alibaba, which adopts the 20B-parameter MM-DiT architecture, achieves state-of-the-art (SOTA) performance across most benchmark evaluations.

## 4.2 Results and Comparisons

In this section, we compare our vision-aligned text feature supervision method against the conventional supervision strategy across two representative SOTA image generation baselines.

As shown in Table 1, our vision-aligned text feature supervision method consistently outperforms both the base models and conventional SFT across all evaluation metrics, indicating that our method successfully facilitates a synergistic enhancement in cross-modal alignment, human-perceived aesthetics, and overall image realism. Our method achieves the highest CLIPScore on both Flux.1-dev and Qwen-Image, validating that incorporating SigLIP 2 features as an auxiliary supervision signal during training effectively injects dense fine-grained cross-modal priors so as to enhance prompt adherence. In particular, compared with the conventional supervision method, our method successfully reduces FID without compromising ImageReward. This is primarily because vision-aligned SigLIP 2 text features are optimized on multi-granularity visual details, inherently encapsulating rich patch-level information.

In Table 2, we compare our method with a well-designed realistic portrait generation model Flux.1-Krea-dev [10]. Experimental results demonstrate that training with manually selected high-quality images introduces a noticeable trade-off: while it effectively improves FID compared to the baseline (35.17 -> 28.20), it comes at the expense of human-perceived aesthetics, leading to a severe degradation in ImageReward (21.32% -> 15.14%). In contrast, our method comprehensively outperforms Flux.1-Krea-dev. By achieving a superior FID of 27.40, our method takes the lead across other key metrics, including CLIPScore (81.40%), HPSv2.1 (30.54%), and ImageReward (21.57%), successfully breaks this trade-off dilemma.

**Table 3: Ablation studies on the architecture and hyperparameters of our feature supervision module.**

| Configuration | FID ↓ | CLIPScore(%) ↑ | HPSv2.1(%) ↑ | ImageReward(%) ↑ |
|---|---|---|---|---|
| Baseline | 35.17 | 79.36 | 30.50 | 21.32 |
| **Supervision VFM** | | | | |
| **SigLIP 2** | 27.40 | **81.40** | **30.54** | **21.57** |
| OpenCLIP | **26.95** | 81.36 | 30.53 | 18.62 |
| **Projection Type** | | | | |
| **MLP** | **27.40** | 81.40 | 30.54 | **21.57** |
| Q-Former | 28.33 | **81.44** | **30.61** | 20.63 |
| **Injection Depth** | | | | |
| **Depth 6** | 27.40 | 81.40 | **30.54** | **21.57** |
| Depth 12 | 27.37 | **81.56** | 30.47 | 19.18 |
| Depth 18 | **26.96** | 81.24 | 30.44 | 19.58 |
| **Loss Weight ($\lambda$)** | | | | |
| $\lambda = 0.1$ | **27.40** | 81.40 | 30.54 | **21.57** |
| $\lambda = 0.2$ | 28.61 | 81.93 | 30.58 | 20.95 |
| $\lambda = 0.3$ | 29.70 | **82.36** | **30.73** | 21.18 |
| $\lambda = 0.4$ | 30.09 | 82.28 | 30.71 | 21.32 |
| $\lambda = 0.5$ | 29.57 | 82.27 | 30.63 | 21.32 |
| $\lambda = 0.6$ | 29.28 | 82.04 | 30.63 | 20.32 |

## 4.3 Ablation Studies

In this section, we discuss the key factors of our proposed training method and provide a detailed analysis of each critical component. Based on the Flux.1-dev, we analyze the effects of four factors on the final generation results: supervision VFM, projection type, injection depth, and weighting coefficients.

*4.3.1 Supervision VFM.* We select two different VFM models acting as teacher networks for feature supervision to analyze how different supervised VFMs affect the training results. To thoroughly validate our method, we compare our default SigLIP 2 (siglip2-so400m) against the massive OpenCLIP (ViT-bigG-14), which features a substantially larger text encoder. Our results reveal that when deploying the large-scale ViT-bigG-14 teacher, the model achieves exceptional distributional realism (FID = 26.95) and improved semantic alignment (CLIPScore = 81.36%) compared with the baseline model, as illustrated in Table 3. However, this comes at the expense of human-perceived aesthetic quality, evidenced by a sharp drop in ImageReward (21.32% -> 18.62%). This observation indicates that traditional global contrastive representations, despite their large parameter size, lack the multi-granularity information necessary to advance aesthetic quality, thus leading to visually unpleasant compositions. In contrast, our lightweight SigLIP 2 teacher overcomes this trade-off. Equipped with dense, patch-level spatial aligned text features inherited from SigLIP 2, it provides a gentle and fine-grained guidance signal. Consequently, it achieves superior text-image alignment (CLIPScore = 81.40%) and highly competitive realism (FID = 27.40), while notably preserving and even increasing ImageReward (21.32% -> 21.57%). This compelling result demonstrates that the superiority of our method stems from its vision-aligned priors, rather than a mere scaling of model parameters.

*4.3.2 Projection Type.* In the FLUX.1-dev architecture, the intermediate unified representation possesses a high dimensionality of 3072. The dimensions do not match those of the teacher model. To compute the feature alignment loss against the teacher model, a projection module is required. In this section we compare two projection types, namely linear MLP and Q-Former blocks. We conduct a detailed comparison for those two projections. As shown in Table 3, while the Q-Former achieves marginally higher scores in basic semantic metrics (e.g., CLIPScore 81.44% and HPS 30.61%), it introduces a degradation in both overall generative realism (FID increases to 28.33) and human-perceived aesthetics (ImageReward drops to 20.63%). The reason is that the Q-Former relies on a fixed number of learnable queries to extract representations, which inevitably compresses and discards the dense and fine-grained image features of MM-DiT. In contrast, a simple point-wise MLP operates as a spatial projection, and seamlessly maps the high-dimensional (3072-d) MM-DiT features into the target spatial dimension while perfectly preserving the exact sequence length. Consequently, the MLP effectively preserves image details and ensures the synergistic enhancement of realism, aesthetics and semantics.

*4.3.3 Injection Depth.* Intuitively, the depth of VFM supervision has a significant impact on the final performance. Thus, we conduct a series of experiments to determine the optimal injection depth. Specifically, inspired by REPA [30], we impose supervision at the 1/3, 2/3, and 3/3 stages of MM-DiT, corresponding to the 6th, 12th, and 18th layers respectively, to thoroughly analyze how injection depth affects the final generative performance.

As shown in Table 3, we investigate the impact of feature injection depth on the MM-DiT architecture. The results reveal a clear trade-off between realism and human-perceived aesthetics. Empirically, shallow layers in MM-DiT govern the global semantic planning and layout of generated images, while deep layers are responsible for high-frequency details and texture rendering. Injecting features at deep layers (e.g., Layer 18) acts as strict detail supervision, achieving the best Fréchet Inception Distance (FID = 26.96). However, such deep-layer intervention forces the model to overly prioritize fine-grained realism, severely degrading the overall aesthetic quality of generated images. By disrupting the inherent aesthetic priors of the base model, it leads to a drop in human aesthetic preference scores (ImageReward falls to 19.58%). In contrast, applying supervision at shallow layers (Layer 6) serves as a gentler guidance. It avoids overwriting the model's native generative manifold, thereby achieving the highest aesthetic quality (ImageReward 21.57%) while maintaining highly competitive FID (27.40), CLIPScore (81.40%) and HPS (30.54%). Since the core objective of our method is to achieve a synergistic balance rather than optimizing a single metric, we empirically select Layer 6 as our default configuration.

*4.3.4 Weighting Coefficients.* As formulated in Equation 6, the hyperparameter $\lambda$ controls the supervision strength of the fine-grained vision-aligned text features. Therefore, we conduct a series of experiments to verify the effect of the value of $\lambda$ on the final generation performance. Hyperparameter $\lambda$ ranges from 0.1 to 0.6 to investigate the sensitivity of the model to the feature supervision.

As illustrated in Table 3, the results clearly demonstrate a trade-off among image realism, text alignment and aesthetic quality.

*Image Realism.* The lowest FID (27.40) is achieved at $\lambda$=0.1, indicating the best image realism. As $\lambda$ increases, FID rises sharply, peaking at 30.09 when $\lambda$=0.4, which suggests that excessive supervision constraints disrupt the native generative manifold and degrade visual fidelity.

*Text-Image Alignment.* Both metrics peak at $\lambda$=0.3 (CLIPScore: 82.36%, HPS: 30.73%), showing that moderate supervision effectively strengthens text-image alignment. However, further increasing $\lambda$ beyond 0.3 leads to a slight decline in both scores, implying that overly rigid constraints do not continuously improve alignment and may introduce unintended artifacts.

*Human Aesthetic Preference.* The highest ImageReward (21.57%) is obtained at $\lambda$=0.1, preserving the model's inherent aesthetic priors. While mid-range $\lambda$ (0.3–0.5) maintains stable aesthetic scores, an extreme weight ($\lambda$=0.6) causes a drop to 20.32%, as rigid enforcement has an adverse effect on the overall visual appeal.

Thus, setting $\lambda = 0.1$ provides a lightweight yet highly effective supervision. It successfully strikes the optimal synergistic balance and achieves Pareto optimization: the absolute best image realism (FID 27.40) and human-perceived aesthetics (ImageReward 21.57%) while maintaining highly competitive prompt adherence.

## 5 Conclusion

This work achieves Pareto improvement in text-to-image diffusion models for human portrait generation, namely the inherent trilemma among text-image alignment, photorealism, and human-perceived aesthetics. Conventional SFT can improve realism but often causes overfitting, corrupts pre-trained visual priors, and degrades either alignment or aesthetics. Preference-based methods such as DPO and RLHF rely on large-scale annotated data or complex reward models, leading to prohibitive costs and limited scalability. To break this zero-sum dilemma, we propose a parameter-efficient, zero-inference-overhead feature supervision paradigm for MM-DiT.

The core idea is to leverage the vision-aligned text features from the advanced vision foundation model SigLIP 2, implicitly distill fine-grained visual semantic priors into the image branch of MM-DiT during training. Using a lightweight lossless MLP projection, we apply feature alignment supervision only at shallow layers to provide precise and gentle guidance without impairing the native generative performance. Unlike traditional pipelines, our method extracts high-quality implicit aesthetic signals directly from fine-grained vision-aligned text features, eliminating the need for preference datasets or reward model training.

Our extensive experiments demonstrate that this method consistently outperforms both base models and SFT versions across key metrics, including FID (photorealism), CLIPScore and HPSv2 (text-image alignment), and ImageReward (human aesthetics). Ablation studies validate the effectiveness of SigLIP 2 as a superior teacher model and further verify that the MLP projection type better preserves fine-grained visual features compared with Q-Former. Moreover, our experiments confirm that shallow-layer supervision maintains natural generalization and aesthetics better than deep-layer intervention.

In summary, this method offers a simple, highly efficient, and low-cost solution to the long-standing trilemma in human portrait generation. While our current framework is primarily validated on the MM-DiT architecture, exploring its generalization to other mainstream diffusion backbones remains a compelling direction for future work. Ultimately, this work presents a reliable supervised training method with significant implications for both academic research and industrial deployment.